\documentclass{article}
\usepackage[nonatbib,final]{neurips_2022}
\usepackage[utf8]{inputenc} 
\usepackage[T1]{fontenc}    
\usepackage{hyperref}       
\usepackage{url}            
\usepackage{booktabs}       
\usepackage{amsfonts}       
\usepackage{nicefrac}       
\usepackage{microtype}      
\usepackage{xcolor}         

\usepackage{amssymb}
\usepackage{amsthm}
\usepackage{amsmath}
\usepackage{svg}
\usepackage{arydshln}
\usepackage{wrapfig}
\usepackage{soul}
\usepackage{subcaption}

\def\LOC{\(\mathbf{LOC}\)}
\def\SSL{\(\mathbf{SSL}\)}
\def\IN{\(\mathbf{IN}\)}
\def\OBJ{\(\mathbf{OBJ}\)}
\def\CLS{\(\texttt{CLS}\)}

\usepackage{todonotes}

\title{Towards Unsupervised Visual Reasoning: \\
Do Off-The-Shelf Features Know How to Reason?}

\author{%
    Monika Wysoczańska$^1$\thanks{Corresponding author: 
    \texttt{monika.wysoczanska.dokt@pw.edu.pl}} ,
   Tom Monnier $^2$,
   Tomasz Trzciński$^{1,3,4,5}$,
   David Picard $^2$ \\
  $^1$Warsaw University of Technology, $^2$LIGM, Ecole des Ponts, Univ Gustave Eiffel\\
  $^3$Tooploox $^4$IDEAS NCBR $^5$Jagiellonian University of Cracow
}

\begin{document}

\maketitle
    
\begin{abstract}
  Recent advances in visual representation learning allowed to build an abundance of powerful off-the-shelf features that are ready-to-use for numerous downstream tasks. This work aims to assess how well these features preserve information about the objects, such as their spatial location, their visual properties and their relative relationships. We propose to do so by evaluating them in the context of visual reasoning, where multiple objects with complex relationships and different attributes are at play. 
  More specifically, we introduce a protocol to evaluate visual representations for the task of Visual Question Answering. 
  In order to decouple visual feature extraction from reasoning, we design a specific attention-based reasoning module which is trained on the frozen visual representations to be evaluated, in a spirit similar to standard feature evaluations relying on shallow networks. 
  We compare two types of visual representations, densely extracted local features and object-centric ones, against the performances of a perfect image representation using ground truth. 
  Our main findings are two-fold. First, despite excellent performances on classical proxy tasks, such representations fall short for solving complex reasoning problem. Second, object-centric features better preserve the critical information necessary to perform visual reasoning. In our proposed framework we show how to methodologically approach this evaluation.
\end{abstract}

\section{Introduction}
\vspace{-5pt}
Visual representation learning has gained a lot of attention thanks to advanced 
frameworks demonstrating unprecedented results without relying on explicit supervision. On the one 
hand, classical self-supervised methods~\cite{wu2018unsupervised, chen2020simple, grill2020bootstrap, caron2021emerging, he2020momentum} producing localized features (i.e., features
that densely correspond to regions of the image)
are exhaustively evaluated on standard tasks like image classification or object detection 
where they perform on par with the supervised ones. On the other hand, more recent 
unsupervised systems~\cite{burgessMONetUnsupervisedScene2019, greffMultiObjectRepresentationLearning2019, locatello2020, monnier2021dtisprites} aiming at learning object-centric representations (i.e.,
each feature is associated with an object in the image) are typically evaluated for 
instance segmentation where benchmarks are saturated.

In this work, we investigate how well off-the-shelf features extract meaningful information about the objects 
in a given image. We propose to evaluate the features' ability to model objects through the performance of a 
reasoning module trained for Visual Question Answering (VQA) by introducing
a new VQA evaluation protocol which is based on a simple attention-based reasoning module 
learned on top of the \emph{frozen} visual features to be evaluated. Similar to feature evaluations that use shallow 
networks for predictions, we aim to decouple 
visual extraction from 
reasoning. 
For a fair comparison, we limit the size of the visual features to a small and fixed number.
Additionally, we investigate the effect of the training set size to discover what is the minimum size required to be able to learn reasoning patterns and if some types of visual representations are better than others at preventing learning spurious correlations.

Our VQA evaluation enables us to compare off-the-shelf features, either densely extracted local features or object-centric representations and to make several findings that are the main contributions of this paper:
First, such representations have excellent performances using classical proxy tasks, their VQA performance in our constrained setup is far from the one attained by the ground truth or the state-of-the-art obtained using dedicated architectures. With this observation, we stress the importance of using complex reasoning tasks, such as VQA, as a complementary evaluation protocol for testing the effectiveness of off-the-shelf image representations.
Second, we find that object-centric features are more suited for visual reasoning than local features. Although this is conceptually expected, we provide an empirical way to exhibit this behaviour. 
Finally, having a limited training set size prohibits learning correct reasoning patterns with all visual representations, although explicit representations seem to be better at preventing learning spurious correlations than implicit ones.

\begin{wrapfigure}{r}{0.6\textwidth}
    \vspace{-40pt}
    \centering
    \includegraphics[width=0.6\textwidth]{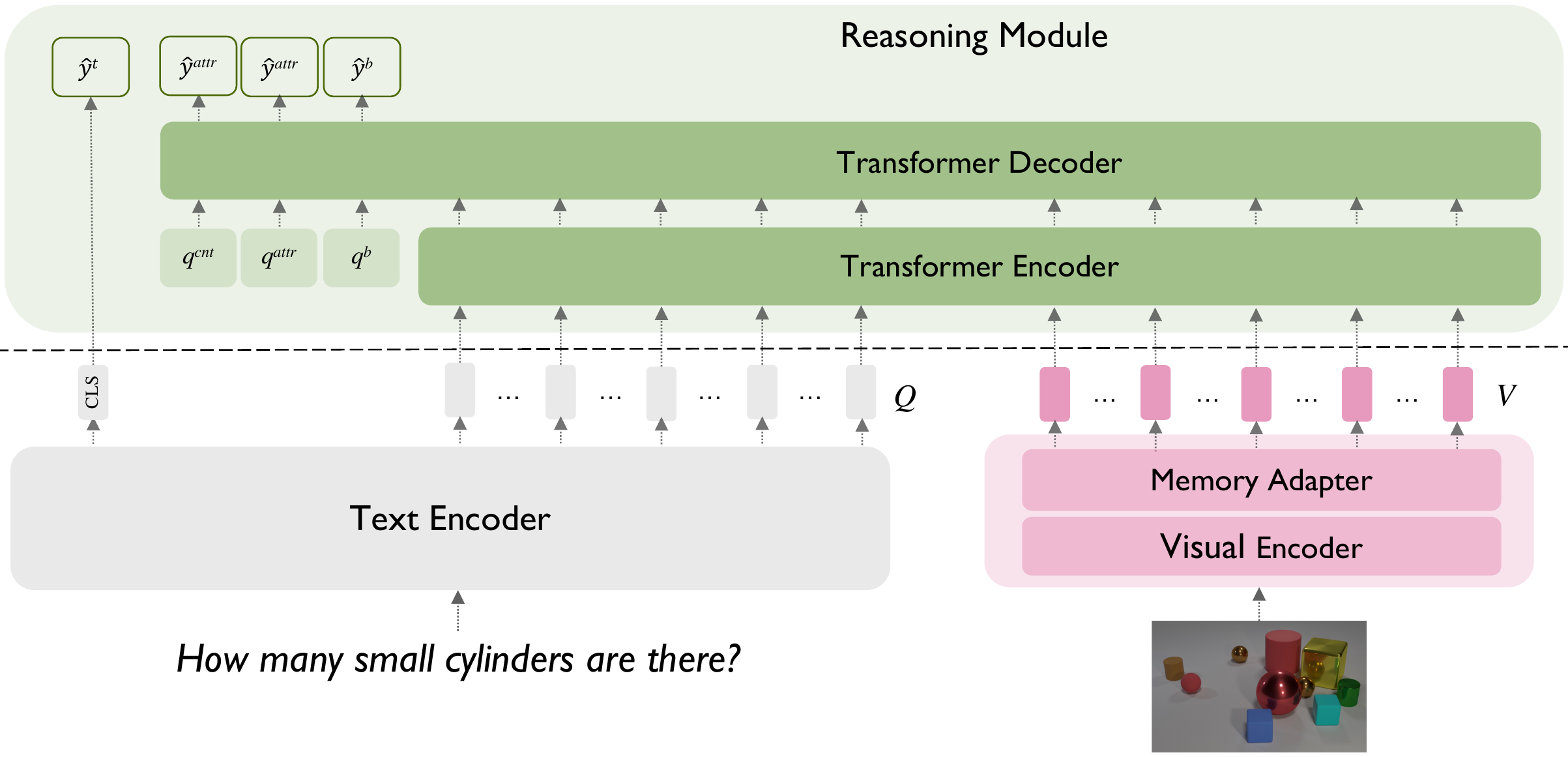}
    \vspace{-15pt}
    \caption{
    Our evaluation framework overview. 
    During training, only the parameters of the reasoning module are trained (above a dashed line), while the rest of the pipeline remains frozen.}
    \vspace{-6pt}
    \label{fig:pipeline}
\end{wrapfigure}

\vspace{-5pt}
\section{Evaluation framework}
\vspace{-6pt}
Our goal is to examine to what extent different off-the-shelf image representations are capable of encoding the information needed for reasoning. 
We examine popular off-the-shelf features that we split into two groups: (i) the classical dense sets of features localized on a grid-like structure, which we refer to as \LOC, and (ii) object-centric features which can be associated with objects in the scene, denoted here as \OBJ. In the case of learned local features, we further distinguish 2 types of learning processes. We examine the standard approach of transferring features obtained from backbones pre-trained for classification on ImageNet~\cite{imagenet}, which we denote by \IN. We also study features obtained through more recent self-supervised learning frameworks, denoted by \SSL.

\vspace{-10pt}
\paragraph{Pipeline overview}

We build our pipeline upon the \textit{disentangling reasoning from the vision and language understanding} paradigm~\cite{Yi2018NeuralSymbolicVD}. Our framework serves as plug-and-play for unimodal encoders and consists of 3 separate modules, for text, vision and reasoning respectively. First, given a question-image pair, we use frozen text and visual encoders to extract features and map them to a common multimodal space. Then, the concatenated features are fed to the reasoning module predicting the answer. An overview of our pipeline is presented in Figure~\ref{fig:pipeline}. 

We use a predetermined and fixed \textbf{text encoder} to ensure a fair comparison of the visual features. Precisely, we use the RoBERTa language model introduced in~\cite{liu2019roberta}, however, we note that this could be any other language model that produces a question representation as a sequence of text tokens.

The \textbf{visual module} maps an input image to a set of visual tokens \textit{V} in a two-step process. First, an image encoder extracts a visual representation which is seen as a sequence of feature tokens. Then, to ensure that all features are given the same complexity, we design a module called \textit{memory adapter} whose goal is to convert the extracted representations to a fixed-size input. 
Inspired by the state-of-the-art model MDETR~\cite{kamath2021mdetr}, our \textbf{reasoning module} is a transformer encoder-decoder~\cite{vaswani2017attention} which operates on the text and visual token sequences \textit{Q} and \textit{V}. To allow our reasoning module to distinguish between modalities, we add to each token a modality-specific segment embedding which is a learnable parameter. In the case of local features, we also add a learned positional encoding to incorporate spatial information, as they do not inherently have such information.

\vspace{-10pt}
\paragraph{Visual memory adaptation}

To fix the input memory size of \textit{V}, we apply a two-step process. First, depending on whether the visual encoder is \OBJ~or \LOC, the sequence length $N_v$ may differ. Thus, we restrict the $N_v$ to be roughly equal to the maximum number of objects in a scene.
Let us assume we have $K$ maximum number of objects that can appear in a scene. Therefore, if $N_v > K$, which is the case for \LOC~due to the grid-like structure of the local features, we apply adaptive average pooling on the features until we roughly match size of $K$. 

Second, to ensure the reasoning module operates on compact representations of similar sizes in memory, we constrain the dimension of the visual tokens. 
Let us assume $d^{min}_{v}$ is the minimum visual token dimension needed to solve the task, which corresponds to a number of objects’ properties as well as their respective positions in a scene for relational reasoning. If $d_{v}$ > $d^{min}_{v}$ we compress visual tokens using Principal Component Analysis (PCA) and decrease the dimensionality to match $d^{min}_{v}$.

Therefore, we define the minimum memory size $M$ required to solve the task as:
$M = K \cdot d^{min}_{v}$.
The memory adapter converts the output of the visual encoder to a fixed memory size within a few orders of magnitude of $M$ by relaxing the $d^{min}_{v}$ constraint since visual features are not expected to attain perfect compression of the visual information. 

\vspace{-10pt}
\paragraph{Training strategy}
We train our reasoning module using question-image-answer triplets $(q, i, a)$ without any external supervision. Concretely, given a triplet we predict the answer type $\hat{y_t}$ as well as the corresponding answer encoded by $\hat{y}_b$, $\hat{y}_{cnt}$ or $\hat{y}_{attr}$ respectively associated to \textit{True/False}, \textit{count} and \textit{attribute}
questions. Our final loss is defined as:
$L_{total} = L_t + L_b + L_{cnt} + L_{attr}$,
 where $L_t$, $L_{cnt}$, $L_{attr}$ denote cross entropy losses between the ground truth $y_t, y_{cnt}, y_{attr}$, and the predictions $\hat{y_t}, \hat{y}_{cnt}, \hat{y}_{attr}$, whereas $L_b$ is a binary cross entropy loss between $y_{b}$ and $\hat{y}_{b}$.
 
\vspace{-5pt}
\section{Experiments}
\begin{wraptable}{r}{0.45\textwidth}
\vspace{-50pt}
\centering
\resizebox{\linewidth}{!}{ 
\begin{tabular}{@{}lccccccc@{}}

\\
\toprule
& method & overall & count & exist & comp num & query attr & comp attr
\\
\midrule
\midrule

\textit{100 mem size} & \\
& GT & 99.7 & 99.8	& 99.9 & 99.5 & 99.2 & 100.0\\

\OBJ & DTI-Sprites~\cite{monnier2021dtisprites} & 81.8 & 72.6 & 87.8 & 84.8 & 84.3 & 82.4 \\
\OBJ & Slot Attention~\cite{locatello2020} & 58.9 & 50.8 & 70.8 &	70.1 & 58.3 & 55.8 \\
\LOC-\SSL & DINO ResNet-50 & 60.4 & 51.0 &	72.6 & 71.4 & 58.7	& 61.2 \\
\LOC-\SSL & DINO ViT & 52.1 & 45.2 & 66.9 & 69.2 & 46.3 & 53.1 \\ 
\LOC-\IN & ResNet-50 & 57.1 & 47.3 & 69.3 &	69.7 &	56.2 & 56.0 \\
\LOC & Raw & 48.9 &	44.9 & 64.1 & 67.9 & 39.8 &	51.1 \\
\midrule
\textit{1000 mem size} & \\
& GT & 99.8 & 100.0 & 100.0 & 99.8	& 99.1 & 100.0 \\
\OBJ & Slot Attention~\cite{locatello2020} & 91.4 & 85.1 &	95.0 &	92.8 & 93.1 &	92.6 \\
\OBJ & DTI-Sprites~\cite{monnier2021dtisprites} 
& 82.2	& 72.7	& 88.1	& 86.2 & 84.6	& 82.9 \\
\LOC-\SSL & DINO ResNet-50 & 82.4 & 71.4 &	88.8 & 82.9 &	85.7 &	84.7 \\
\LOC-\SSL & DINO ViT & 78.4 & 66.0 & 86.5 &	79.1 & 80.7 & 81.0 \\
\LOC-\IN & ResNet-50 & 79.0 & 66.6 &	85.9 &	81.5 &	83.0 & 80.2 \\
\LOC  & Raw & 50.5 & 44.5 &	64.8 &	69.0 &	43.9 &	51.2 \\
\bottomrule
& 
MDETR~\cite{kamath2021mdetr} & 99.7 & 99.3 & 99.9 & 99.4 & 99.9 & 99.9 
\end{tabular}
} 
\caption{Results on the CLEVR dataset. We report scores on the validation set with a detailed split by the question type. Overall, even in the larger memory setup, there is a significant gap between off-the-shelf features and the state of the art MDETR~\cite{kamath2021mdetr}, which performs on par with ground truth.
 } 
\label{tab:results}
\vspace{-2pt}
\end{wraptable}

\vspace{-5pt}
\subsection{Implementation details}
\vspace{-5pt}
For our comparative study, we choose popular image representations trained with various levels of supervision. Models are always frozen during training and used solely as feature extractors.
We use features extracted from two popular architectures: convolutional-based ResNet-50~\cite{he2016deep} and the transformer-based ViT-S proposed in~\cite{dosovitskiy2020vit}.
To study the influence of the level of supervision in image backbones, in the case of ResNet-50 we evaluate the local features trained in a supervised manner using the classification task on Imagenet, denoted \LOC-\IN{}, and self-supervised DINO features, denoted \LOC-\SSL. For ViT-S, we use features trained with DINO.
To evaluate the performance of~\OBJ~we consider two methods: Slot Attention~\cite{locatello2020} and DTI-Sprites~\cite{monnier2021dtisprites} which demonstrated state-of-the-art segmentation results on the recent CLEVRTex benchmark~\cite{karazija2021clevrtex}. 

We conduct our analysis on the CLEVR dataset~\cite{johnson2017clevr} - a common synthetic benchmark for Visual Question Answering. CLEVR world consists of 3D objects in different shapes, colours, sizes and materials. 
The maximum number of objects in scene \textit{K} is 10. Therefore, the minimum memory size of visual tokens $M = 10 \times 7$, for 10 objects, 4 properties and 3D position (x,y,z).

Since all considered representations are much larger than $M$, we consider 2 memory sizes in our protocol: (i) a total memory of size 100, called 100 mem size, which approximates the min $M$ for CLEVR dataset, and (ii) a total memory of size 1000, called 1000 mem size, to account for richer representations than the bare minimum required for solving the VQA task. 

All the details on adaptation and implementations can be found in Appendix~\ref{sec:ve_details}.

\vspace{-5pt}
\subsection{Results and discussion}
\begin{wrapfigure}{r}{.39\textwidth}
    \vspace{-10pt}
    \begin{subfigure}{.39\textwidth}
        \centering
        \includegraphics[width=\textwidth]{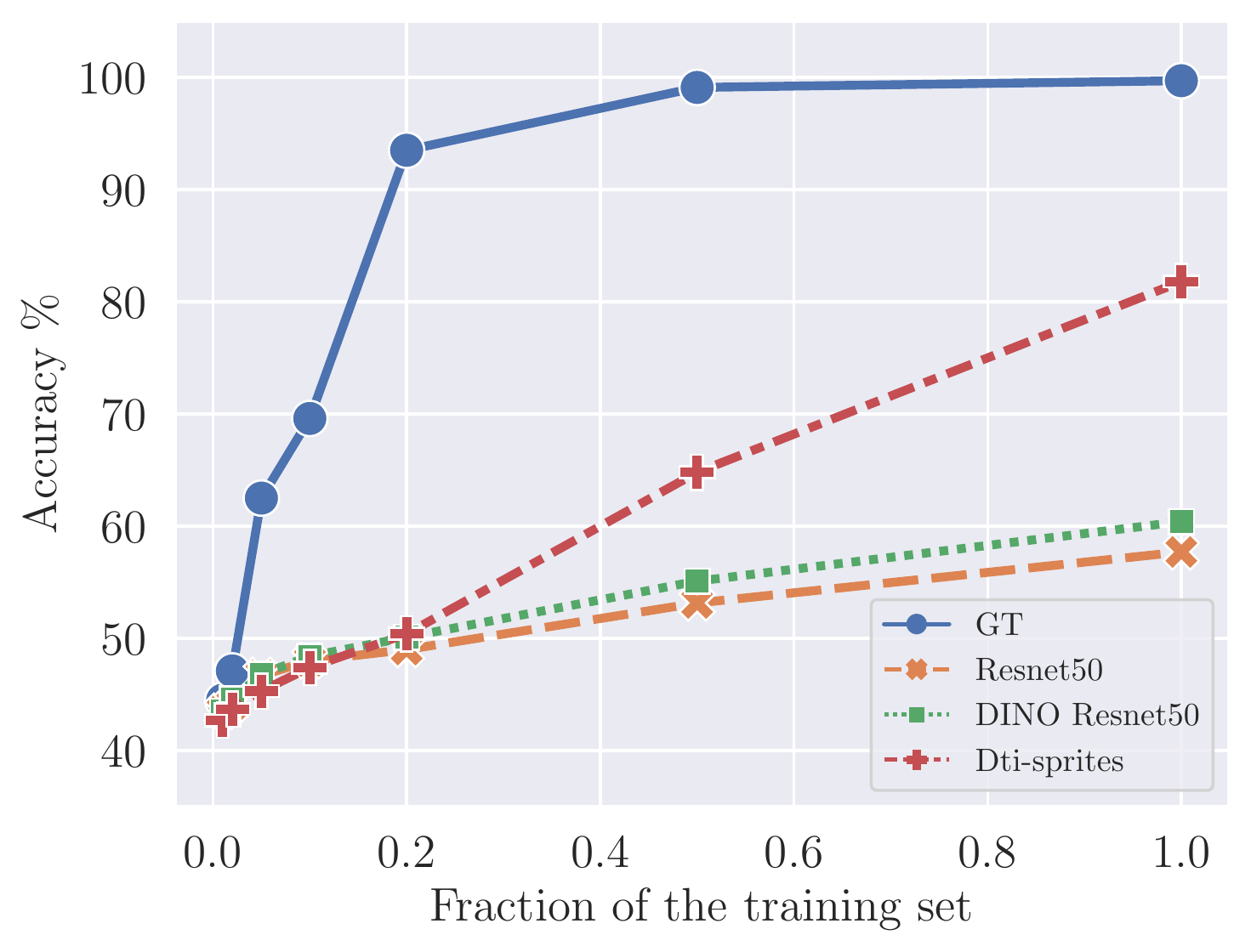}
        \caption{100 mem size}
        \label{fig:fs_100}
    \end{subfigure}
    \begin{subfigure}{.39\textwidth}
        \centering
        \includegraphics[width=\textwidth]{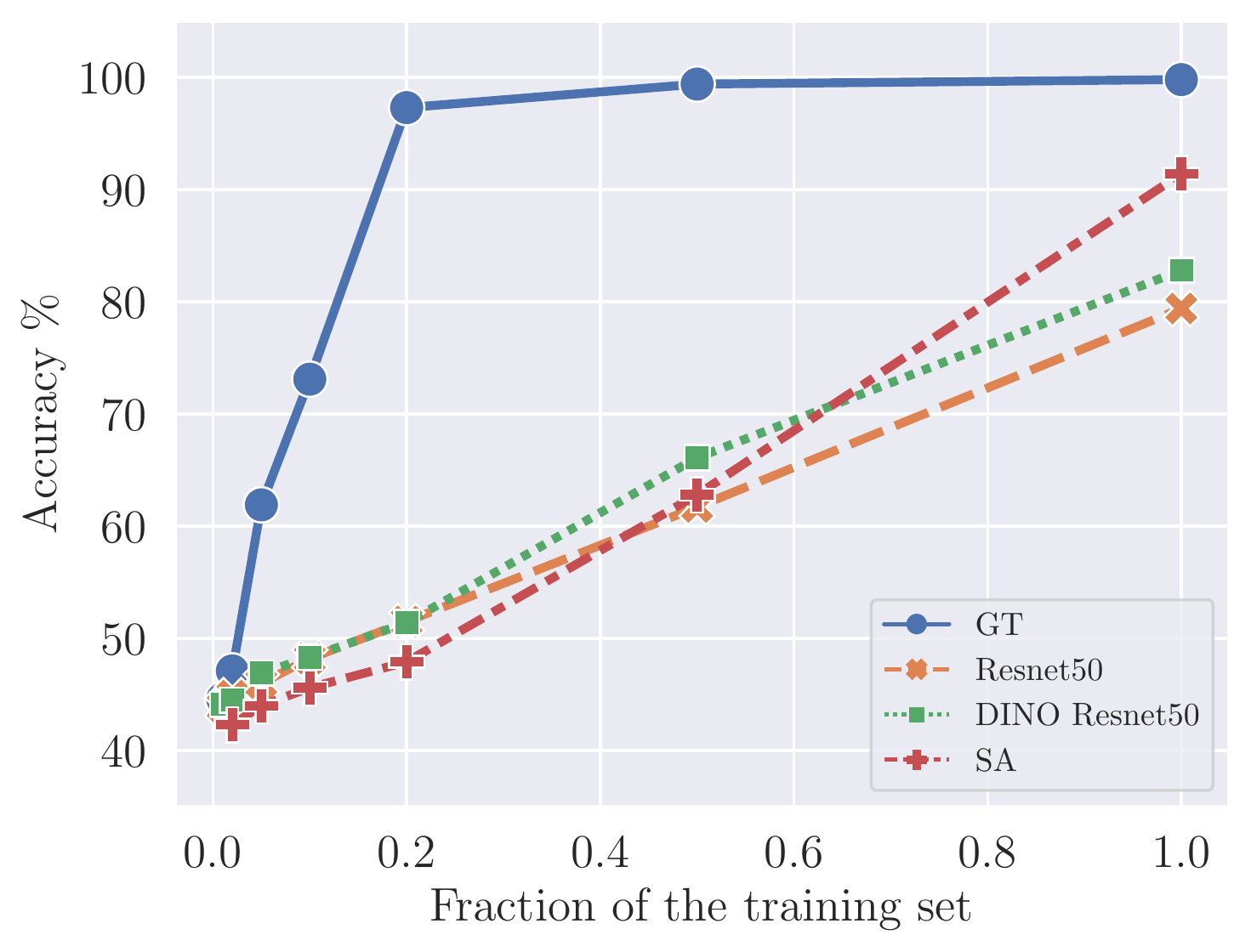}
        \caption{1000 mem size}
        \label{fig:fs_1k}
    \end{subfigure}
    \caption{Results of the few-shot VQA experiments on CLEVR dataset. 
    }
    \vspace{-14pt}
    \label{fig:few_shot}
\end{wrapfigure}

\vspace{-5pt}
We conduct 2 experiments. In the memory size constraint test, we study the effectiveness of different visual features in solving the VQA task when heavily constraining the input size of the reasoning module. The overall accuracy as well as all the question-specific categorical accuracy are shown in Table~\ref{tab:results}. 
Moreover, we study the effectiveness of visual features in solving VQA with a limited number of samples. The idea behind this experiment is to test the generalization capacity of the features as we argue that accurate and discriminative features should generalize better to unseen samples.
Figure~\ref{fig:few_shot} shows a comparison for varying fractions of the training set for both memory sizes. 

\vspace{-6pt}
\paragraph{Are visual representations anywhere close to ground truth performances?}
From the 100 mem size setup, it is clear that all of the methods except for DTI-Sprites fail to attain accuracy which further suggests that they do not encode visual information in a compact way that is comparable to perfect visual information. 
We attribute the overall best performance of DTI-Sprites in this restrictive memory size regime to its inherent explicit representation nature. The original memory size of a DTI-Sprites visual token is the closest to the ground truth, which plays in its favour in this experiment.

\paragraph{Do visual representations contain sufficient information for solving VQA?}
In the more relaxed memory constraint setup (1000 mem size), all the studied visual representations enable the reasoning module to learn to solve VQA to some extent, with Slot Attention clearly outperforming the rest. This may suggest that even if the visual information is not encoded in a way that can be heavily compressed, these features nonetheless contain significant semantic information. 
To study the effect of introducing memory constraint, we also train our reasoning module using original feature sizes. In the case of DINO ResNet-50 features, we obtained 87.9\% overall accuracy, which indicates that our memory adapter does introduce a bottleneck. Nevertheless, features at their best are still far from the ground truth performance. 

\vspace{-6pt}
\paragraph{Are object-centric representations more suitable for the task?}
For the \textit{exist} questions, \OBJ{} features are outperforming \LOC{} ones. This is expected and quantitatively indicates that structuring the scene into a set of objects is much more suitable for reasoning.
When it comes to \textit{comp num, query attr} and \textit{comp attr}, Slot Attention performs significantly better compared to other methods. We argue this can be attributed to the object-centric nature of the representation that facilitates comparisons among objects and focuses on describing their properties. We note that \OBJ indeed were trained on CLEVR contrary to \LOC features. However, we tried fine-tuning DINO features on CLEVR dataset, but we obtained worse results. We hypothesize they are not suited for synthetic datasets like CLEVR, which is smaller and much less diverse than standard SSL datasets. 

\vspace{-6pt}
\paragraph{Are visual representations able to reason from a few examples?}
We observe that starting at 20\% of the full training set and with perfect visual information corresponding to the ground truth, it is possible to infer reasoning patterns. However, at this training set size, none of the considered visual representations enables visual reasoning, regardless of the memory constraint.

\vspace{-6pt}
\paragraph{Are compact and explicit visual representations better suited for learning to reason from a few examples?}
Looking at the 100 mem size setup (Figure \ref{fig:fs_100}), we can see that DTI-Sprites is able to obtain higher accuracy with much fewer examples than the other methods, 
whereas all achieve comparable very low training losses (see ~\ref{sec:fs_det}). This may indicate that explicit representations enable a quicker discovery of relevant information than implicit representations.

\vspace{-6pt}
\paragraph{Do object-centric representations prevent learning spurious correlation?}
Figure \ref{fig:fs_1k} suggests that both \OBJ{} and \LOC{} representations exhibit similar behaviour when the memory is less constrained. Given that they all reach similar very high training accuracy, this indicates that structuring the representation into objects may not be as good at preventing learning spurious correlations as is the distinction between explicit and implicit representations.

\vspace{-5pt}
\section{Conclusions}

\vspace{-6pt}
We investigated to what extent off-the-shelf representations model the information necessary to perform visual reasoning. To that end, we design a new feature evaluation protocol based on VQA which aims at disentangling as much as possible the vision from the reasoning part. 
Using our evaluation protocol, we make three key findings: (i) off-the-shelf visual representations are far from being able to structure visual information in a compact manner, (ii) object-centric representations seem to be better at encoding the critical information necessary for reasoning, and (iii) limiting the training set size has a dramatic impact on the learning of spurious correlations. 
While these findings contrast with the excellent performances that off-the-shelf features usually obtain in simpler vision tasks, they also show that having representations that encode object properties is a promising first step towards unsupervised visual reasoning.

\begin{ack}
This work was supported by HPC resources from GENCI-IDRIS (2021-AD011013085), by Foundation for Polish Science (grant no POIR.04.04.00-00-14DE/18-00) carried out within the Team-Net program co-financed by the European Union under the European Regional Development Fund, as well as the National Centre of Science (Poland) Grant No. 2020/39/B/ST6/01511.
\end{ack}

\bibliography{bib}
\bibliographystyle{unsrt}
\newpage

\section{Appendix}
\subsection{Visual encoders extraction and adaptation details}
\label{sec:ve_details}

\subsubsection{Dense local features} 
We use features extracted from two popular architectures: convolutional-based ResNet-50~\cite{he2016deep} and the transformer-based ViT-S proposed in~\cite{dosovitskiy2020vit}. For ResNet-50, we use the local features after the last convolutional layer right before the global average pooling, whereas for ViT-S, we use the output tokens of the last layer corresponding to the image patch position (\textit{i.e.}, the \CLS~token is not used). To encode the position of local features, we use the corresponding 2D position in the feature map for ResNet-50 features and the 2D position of the corresponding patch for ViT-S features.

For ResNet-50 pre-trained on ImageNet we use model weights available in torchvision package\footnote{\url{http://pytorch.org/vision/stable/index.html}}. For both DINO ResNet-50 and DINO ViT-S/16 we use weights provided in original DINO repository\footnote{\url{http://github.com/facebookresearch/dino}}. 

\subsubsection{Object-centric representations}
To evaluate the performance of unsupervised object-centric representations, we consider two methods: Slot Attention~\cite{locatello2020} and DTI-Sprites~\cite{monnier2021dtisprites} which demonstrated state-of-the-art segmentation results on the recent CLEVRTex benchmark~\cite{karazija2021clevrtex}. We first train both methods on CLEVR dataset - which does not require any supervision - and use pre-trained models as feature extractors. 

\paragraph{Slot Attention}
In Slot Attention we use the slots right before the decoder part as features. We use the implementation available in CLEVRTex benchmark repository \footnote{\url{http://github.com/karazijal/clevrtex}}. We train the model on the original, CLEVR VQA dataset, preserving the original train/validation splits. To keep a similar resolution as in the case of the multi-object segmentation task, we feed input images resized to $120\times 160$. We do not apply any centre cropping to make sure all the objects in the scenes are clearly visible.
Following~\cite{karazija2021clevrtex} we use 11 slots, we also maintain the original learning
rate, batch size, and optimizer settings as well as 500k iterations of training. 

\paragraph{DTI-Sprites}
We use original implementation of DTI-Sprites \footnote{\url{http://github.com/monniert/dti-sprites}}. Similar to Slot Attention, we train the model on images resized to $120\times 160$ with no centre cropping. To account for smaller resolution compared to the original multi-object segmentation task, we increase the representation expressiveness in the backbone by changing adaptive average pooling to $4\times4$, instead of the originally proposed $2\times2$. We train the model with 10 layers, corresponding to a maximum number of objects in the CLEVR dataset. We also increase the number of prototypes to 8 since we observed that the training did not lead to obtaining a complete set of prototypes in the dataset when using only 6. We train the model with the original batch size and learning rate for 760k iterations.

\paragraph{Memory adaptation details}
To match the memory constraints in 2 setups for all visual encoders we use PCA implementation in Scikit-learn~\cite{scikit-learn}. We first extract features for train and validation sets and apply PCA offline. 
\begin{table}
\centering
     \caption{Memory size constraints imposed at the feature adaptation step. 
     We conduct our studies in 2 memory size regimes: \textit{100 mem size}, and \textit{1000 mem} size. We also provide the raw output sizes for each visual encoder to highlight the extent of compression and expansion applied at the adaptation stage.}
\resizebox
{\linewidth}{!}{ 

    \begin{tabular}{l|cccc}
        \toprule
        method & input image size & raw output size & 100 mem size & 1000 mem size \\
        \midrule
        \midrule
        GT & - & 10$\times$7 & 10$\times$10 & 10$\times$100 \\
        DTI-Sprites & 120$\times$160  & 10$\times$10 & 10$\times$10 & 10$\times$100 \\
        Slot Attention & 120$\times$160 & 11$\times$64 & 11$\times$9 & 11$\times$90\\
        ResNet-50 & 224$\times$224 & 49$\times$2048 & 16$\times$6 & 16$\times$62 \\
        ViT-S & 224$\times$224 & 194$\times$384 & 16$\times$6 & 16$\times$62 \\
        Raw & 192$\times$192 & 9$\times$12k & 9$\times$11 & 9$\times$111  
    \end{tabular}
}

\label{tab:regimes}
\end{table}

\subsection{Training details}
\label{sec:train_det}

We implement our framework in PyTorch~\cite{pytorch}. For the text encoder we use RoBERTa\footnote{\url{http://huggingface.co/roberta-base}} model available in HuggingFace library~\cite{wolf2019huggingface}. In the case of all visual encoders, we follow the same training strategy. We train the reasoning module for 40 epochs, with a batch size of 64. We use adamW~\cite{loshchilov2017decoupled} optimizer with a learning rate of 1e-4, weight decay of 1e-4 and linear warmup for the first 10k iterations. We then decrease the learning rate at epochs 30th and 35th by a factor of 10.

\paragraph{Computational cost} Regardless of the visual encoder used as an input, a full training run takes approximately 20 hours on a single NVIDIA Tesla V100 GPU. In total we used approximately 15k GPU hours for both obtaining visual features as well as for the evaluation phase.

\subsection{Details on the training set size study}
\label{sec:fs_det}

\begin{figure}[h]
\centering
    \begin{subfigure}{.78\textwidth}
        \centering
        \includegraphics[width=\textwidth]{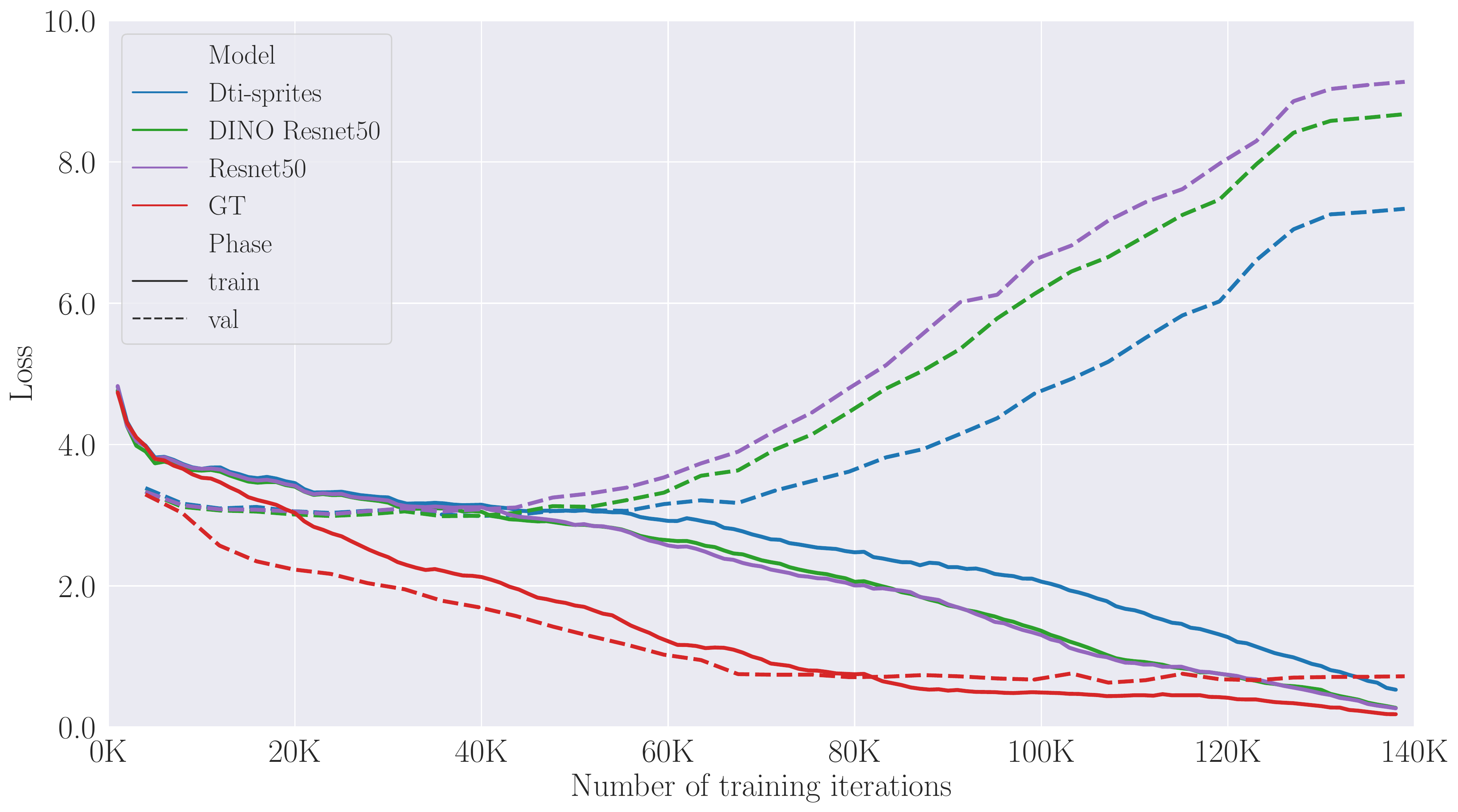}
        \caption{100 mem size}
        \label{fig:fs20_100}
    \end{subfigure}
    \begin{subfigure}{.78\textwidth}
        \centering
        \includegraphics[width=\textwidth]{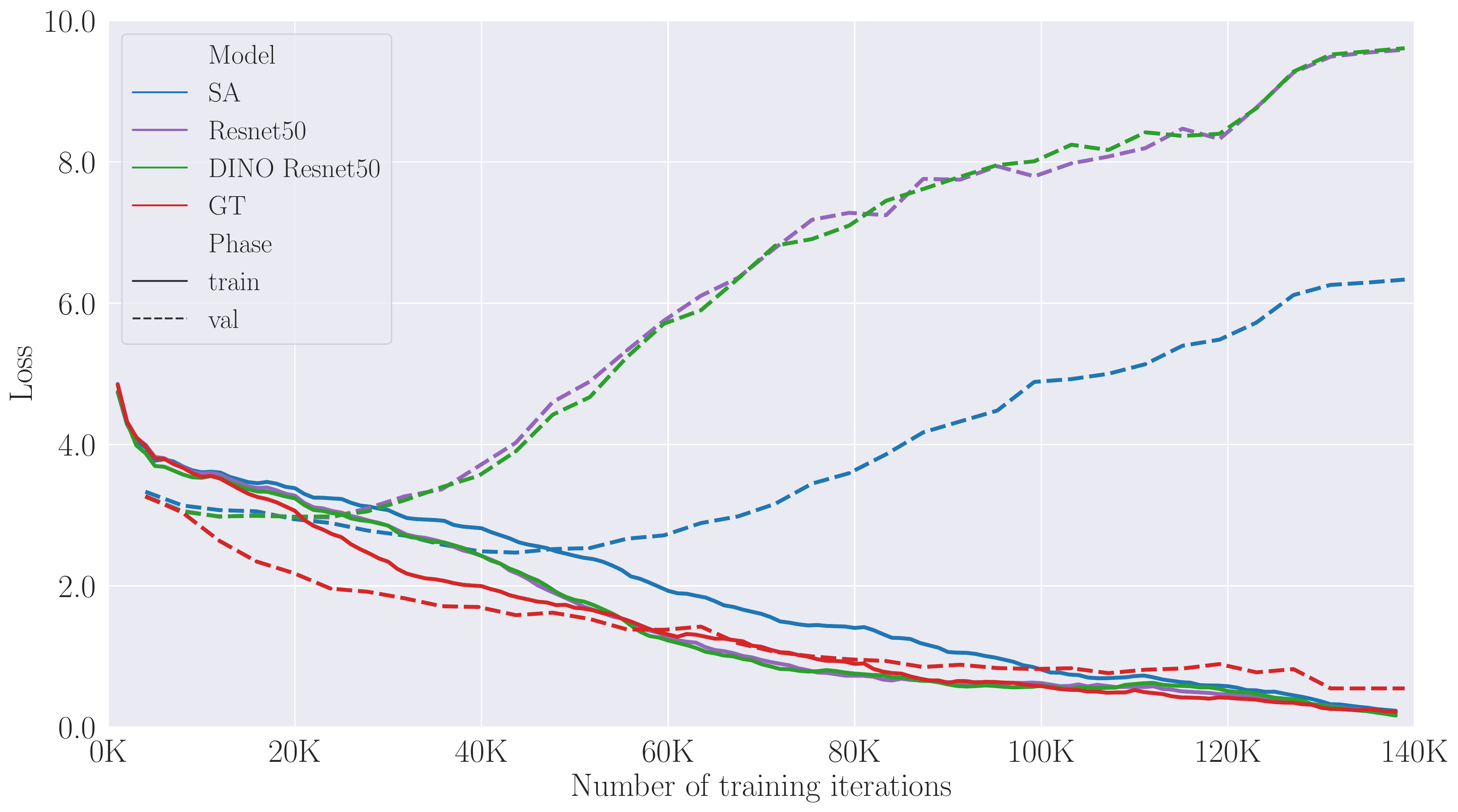}
        \caption{1000 mem size}
        \label{fig:fs20_1k}
    \end{subfigure}
    \caption{Training and validation curves at a 20\% fraction of the training set. We report training loss for 20\% of the training set being used, while validation loss we calculate over the whole validation set in the CLEVR dataset.}
    \label{fig:20_lr_curves}
\end{figure}

In the training set study, we limit the number of training samples by reducing the number of scenes. We observe that starting at 20\% of the full training set with perfect visual information corresponding to the ground truth, it is possible to obtain high validation accuracy. Therefore, we study the effectiveness of visual encoders at this threshold. 

Figure~\ref{fig:20_lr_curves} shows the learning curves at a 20\% fraction of the training set for both memory size regimes. In Figure~\ref{fig:fs20_100}, which depicts the learning curves in 100 mem size, in the case of all of the methods we observe that the model starts to quickly overfit, with DTI-Sprites performing relatively better. 

In the case of 1000 mem, Figure~\ref{fig:fs20_1k} size we observe the overfitting effect happening even earlier in the training process. This may indicate the occurrence of leakage from visual encoders to the reasoning part. 

In both memory size regimes \OBJ{} representations demonstrate relatively better generalization capabilities but are still far from the ground truth information.


\end{document}